\documentclass[conference,letterpaper]{IEEEtran}
\usepackage[letterpaper, left=0.7in, right=0.7in, bottom=0.99in, top=0.7in]{geometry}
\IEEEoverridecommandlockouts

\usepackage{cite}
\usepackage{amsmath,amssymb,amsfonts, mathtools, amsthm}
\usepackage{algorithmic}
\usepackage{algorithm}
\usepackage{lipsum}
\usepackage{subcaption}
\usepackage{epsfig}
\usepackage{bbm}
\usepackage{bm}
\usepackage{booktabs}
\usepackage{multirow}
\usepackage[normalem]{ulem}
\usepackage{graphicx, subfig}
\usepackage{textcomp}
\usepackage{xcolor}
\usepackage{verbatim}
\usepackage{glossaries}
\usepackage{threeparttable}
\usepackage{tabularx}
\usepackage{caption}
\usepackage{dsfont}
\usepackage{titlesec}
\usepackage[bb=dsserif]{mathalpha}
\usepackage{pgfplots}

\newacronym{bcd}{BCD}{Block Coordinate Descent}
\newacronym{leo}{LEO}{Low Earth Orbit}
\newacronym{isl}{ISL}{Inter-Satellite Link}
\newacronym{gsd}{GSD}{ground sample distance}
\newacronym{fov}{FoV}{field of view}
\newacronym{gtfp}{GTFP}{ground track frame period}
\newacronym{gs}{GS}{ground station}
\newacronym{fso}{FSO}{free-space optical}
\newacronym{smec}{SMEC}{satellite mobile edge computing}
\newacronym{mec}{MEC}{mobile edge computing}
\newacronym{pdd}{PDD}{Penalty Dual Decomposition}
\newacronym{aodv}{AODV}{Ad hoc On-demand Distance Vector}
\newacronym{cnn}{CNN}{convolutional neural networks}
\newacronym{dl}{DL}{Deep Learning}
\newacronym{dod}{DoD}{depth of discharge}
\newacronym{dqn}{DQN}{deep Q-learning}
\newacronym{gsl}{GSL}{ground-to-satellite link}
\newacronym{bm}{BM}{benchmark}
\newacronym{ml}{ML}{Machine Learning}
\newacronym{mdp}{MDP}{Markov decision process}
\newacronym{ngeo}{NGEO}{Non-geostationary orbit}
\newacronym{olsr}{OLSR}{optimized link state routing protocol}
\newacronym{ospf}{OSPF}{Open Shortest Path First}
\newacronym{pan}{PAN}{Path-Aware Networking}
\newacronym{qos}{QoS}{Quality of Service}
\newacronym{rl}{RL}{Reinforcement Learning}
\newacronym{drl}{DRL}{Deep Reinforcement Learning}
\newacronym{dnn}{DNN}{Deep Neural Network}
\newacronym{dql}{DQL}{Deep Q-learning}
\newacronym{e2e}{E2E}{end-to-end}
\newacronym{bgp}{BGP}{Border Gateway Protocol}
\newacronym{ibgp}{iBGP}{interior Border Gateway Protocol}
\newacronym{ebgp}{eBGP}{exterior Border Gateway Protocol}
\newacronym{as}{AS}{Autonomous System}
\newacronym{relu}{ReLu}{Rectified Linear Unit}
\newacronym{cdf}{CDF}{Cumulative Distribution Function}
\newacronym{ntn}{NTN}{Non-Terrestrial Networks}
\newacronym{lsatc}{LSatC}{\gls{leo} Satellite Constellation}
\newacronym{ai}{AI}{Artifical Intelligence}
\newacronym{ip}{IP}{Internet Protocol}
\newacronym{ue}{UE}{User Equipment}
\newacronym{pomdp}{POMDP}{Partially Observable Markov Decision Problem}
\newacronym{hol}{HOL}{Head of Line}
\newacronym{fifo}{FIFO}{First-In First-Out}
\newacronym{snr}{SNR}{Signal-to-Noise Ratio}
\newacronym{eo}{EO}{Earth Observation}
\newacronym{aoi}{AoI}{Age of Information}
\newacronym{paoi}{PAoI}{Peak Age of Information}
\newacronym{semcom}{SemCom}{Semantic Communications}
\newacronym{go}{GO}{Goal-Oriented Communications}
\newacronym{vtw}{VTW}{Visible Time Windows}
\newacronym{otw}{OTW}{Observation Time Windows}
\newacronym{aeossp}{AEOSSP}{Agile Earth Observation Satellite Scheduling Problem}
\newacronym{aeos}{AEOS}{Agile Earth Observation Satellites}
\newacronym{ceos}{CEOS}{Conventional Earth Observation Satellites}
\newacronym{ec}{EC}{Edge Computing}
\newacronym{gcn}{GCN}{Graph Convolutional Networks}
\newacronym{gnn}{GNN}{Graph Neural Networks}
\newacronym{gat}{GAT}{Graph Attention Networks}
\newacronym{sgd}{SGD}{stochastic gradient descent}

\DeclareMathOperator*{\argmax}{arg\,max}

\def\BibTeX{{\rm B\kern-.05em{\sc i\kern-.025em b}\kern-.08em
    T\kern-.1667em\lower.7ex\hbox{E}\kern-.125emX}}

\title{An energy-efficient learning solution for the Agile Earth Observation Satellite Scheduling Problem}

\author{\IEEEauthorblockN{Antonio M. Mercado-Martínez, Beatriz Soret~\IEEEmembership{Senior Member,~IEEE}, Antonio Jurado-Navas~\IEEEmembership{Member,~IEEE}}
\vspace{-0.4cm}

\thanks{This work is funded by the Spanish Ministerio de Ciencia, Innovación y Universidades (project "TATOOINE", grant no. PID2022-136269OB-I00). The author thankfully acknowledges the computer resources, technical expertise and assistance 
provided by the SCBI (Supercomputing and Bioinformatics) center of the University of Malaga.} }

\def\subparagraph{} 

\titlespacing*{\section}{0pt}{*1}{*1}
\titlespacing{\subsection}{0pt}{*1}{*1}

\renewcommand{\thesubsubsection}{\arabic{subsubsection}}

\titleformat{\subsubsection}[runin]{\itshape}{\thesubsubsection)}{1em}{}
\titlespacing*{\subsubsection}{\parindent}{0pt}{*1}

\begin{document}

\bstctlcite{IEEEexample:BSTcontrol}

\maketitle
\begin{abstract}
The \gls{aeossp} entails finding the subset of observation targets to be scheduled along the satellite's orbit while meeting operational constraints of time, energy and memory. The problem of deciding  \emph{what} and \emph{when} to observe is inherently complex, and becomes even more challenging when considering several issues that compromise the quality of the captured images, such as cloud occlusion, atmospheric turbulence, and image resolution. This paper presents a \gls{drl} approach for addressing the \gls{aeossp} with time-dependent profits, integrating these three factors to optimize the use of energy and memory resources. The proposed method involves a dual decision-making process: selecting the sequence of targets and determining the optimal observation time for each. Our results demonstrate that the proposed algorithm reduces the capture of images that fail to meet quality requirements by $> 60\%$ and consequently decreases energy waste from attitude maneuvers by up to $78\%$, all while maintaining strong observation performance.
\end{abstract}

\glsresetall

\section{Introduction}

One of the most relevant advances in the realm of \gls{eo} has been the introduction of \gls{aeos} \cite{Wang_2021}. Unlike \gls{ceos}, which can only adjust their attitude along the roll axis, \gls{aeos} have a strong attitude adjustment capability along three axes (roll, pitch, and yaw). This allows \gls{aeos} to have longer \gls{vtw}, enabling multiple \gls{otw} for a single target. The \gls{vtw} represents the time interval in which a concrete observation target can be observed, while the \gls{otw} represents its real observation time. Although this provides a wide range of options, it also complicates the selection of the \gls{otw} when multiple observation targets are involved in the same observation period, which is known as the \gls{aeossp}. In particular, the \gls{aeossp} consists of maximizing the total observation profit while satisfying all temporal, energy, and memory constraints. Such observation profit can be defined in different ways depending on the nature of the application.

However, this process is not always straightforward, as various factors impact the quality and, consequently, the usability of the collected images. For example, a substantial number of collected \gls{eo} images are discarded because of  cloud occlusion \cite{cloud}. Likewise, the presence of turbulence in the atmosphere causes aberrations in the image during its capture \cite{andrews2005laser}, which also results in its exclusion. Finally, image resolution must be also considered as some applications require high-resolution images to perform properly. This is defined by the \gls{gsd}, i.e., the real distance represented by a pixel in the image, which increases as the target moves away from the nadir. The smaller the \gls{gsd}, the higher the image quality \cite{ijgi10060406}. Nevertheless, the scope of \gls{aeos} allows us to integrate these factors into the \gls{aeossp}, improving the performance and the obtained observation profit.

In this paper, we investigate a \gls{drl} \cite{Sutton1998} approach for solving the \gls{aeossp} with time-dependent profits for a single \gls{aeos} taking into account the presence of clouds and atmospheric turbulence, as well as image resolution, in order to achieve both good energy and memory resources management. This involves a dual decision-making process, as it requires selecting both the sequence of targets and the best observation time for each. To address it, we model the problem as a graph and leverage the use of \gls{gnn}.

The rest of the paper is organized as follows. Section \ref{sec:soa} provides an overview of related works. In Section \ref{sec:sysmodel}, we introduce the system model. The optimization problem is described in Section \ref{sec:aeossp}, while the proposed algorithm is presented in Section \ref{sec:algorithm}. The conducted experiments and the obtained results are discussed in Section \ref{sec:results},  whereas \mbox{Section \ref{sec:conclusions}} concludes the paper.

\section{State of the Art} \label{sec:soa} 

The \gls{aeossp} has been widely studied in recent years. Existing approaches can be classified into exact methods, heuristics, metaheuristics, and machine learning. Among machine learning methods, some studies have employed \gls{drl}, such as in the case of \cite{math11194059}, where \gls{gnn} were applied with promising results. This approach is particularly well suited to the problem, as it can be effectively  modeled  as a graph, a quite common practice in the field \cite{alma991001513474404886}.

Cloud coverage remains a significant challenge within the scope of \gls{aeossp} \cite{alma991001513474404886}.  Various studies have addressed this issue, such as \cite{cloud}, where a re-scheduling algorithm based on cloud coverage forecasting was developed. On the other hand, the presence of atmospheric turbulence has not been as extensively explored in the \gls{aeossp} context. However,  \cite{soret2024semanticgoalorientededgecomputing} evaluates its impact on \gls{eo} images  and concludes that it is an important factor to consider. 

Finally, image resolution is another critical aspect affecting the quality of  \gls{eo} images. This factor was taken into account in \cite{PENG201984}, where an exact method for solving the \gls{aeossp} with time-dependent profits was developed. There, the observation profit was formulated as a function of image resolution. Similar to our approach, this study incorporates a dual decision-making process that involves selecting the sequence of targets and determining the observation time for each.

\section{System Model} \label{sec:sysmodel}
\begin{figure} [t]
\centering
\includegraphics[width=0.43\textwidth]{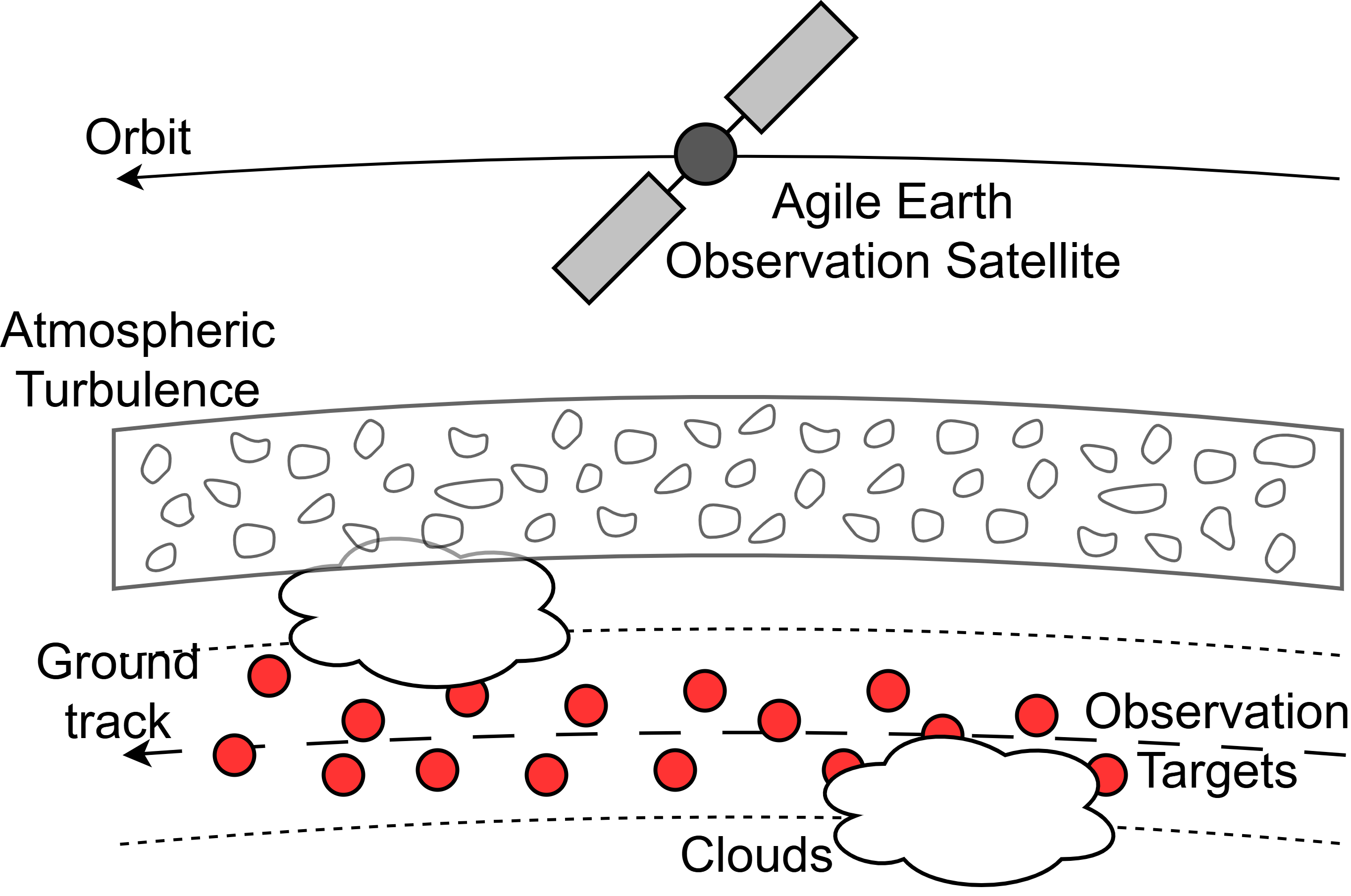}
\caption{Scenario with a single \gls{aeos}, multiple observation targets and the presence of clouds and atmospheric turbulence.}
\label{fig:scenario} \vspace{-0.4cm}
\end{figure}

Fig. \ref{fig:scenario} shows the considered scenario consisting of an \gls{aeos} that observes a set of $N$ ground observation targets along its orbit.  
The satellite operates at an altitude $h_{sat}$ in an orbit with an inclination $\zeta$, observing a set of targets $T$ distributed within an observation area. Each target $i$ represents a squared area of size $l_{target}$, which can be observed at $t_{i,n}$.

The attitude of the satellite during observation at $t_{i,n}$ is defined by the roll ($\theta_{i,n}$), pitch ($\phi_{i,n}$), and yaw ($\psi_{i,n}$) angles. The satellite's attitude maneuver capability  is determined by its maximum roll, pitch, and yaw angles ($\theta_{max}$, $\phi_{max}$, and $\psi_{max}$, respectively) as well as the attitude transition time. These factors dictate which targets can be observed and scheduled at any given time. The attitude transition time is described by a piecewise linear function that depends on the speed of the camera and the angular difference between two consecutive observations \cite{LIU201741}. Specifically, for the scenario where target $i$ is observed at $t_{i, n}$ and target $j$ is observed at $t_{j, m}$, the attitude transition time is defined as follows:
\begin{equation}
    \Delta t_{i,n-j,m} = 
    \begin{cases}
        11.66, & \alpha_{i,n-j,m} \leq 10^\circ \\
        5 + \alpha_{i,n-j,m} / 1.5, & 10^\circ < \alpha_{i,n-j,m} \leq 30^\circ \\
        10 + \alpha_{i,n-j,m} / 2, & 30^\circ < \alpha_{i,n-j,m} \leq 60^\circ \\
        16 + \alpha_{i,n-j,m} / 2.5, & 60^\circ < \alpha_{i,n-j,m} \leq 90^\circ \\
        22 + \alpha_{i,n-j,m} / 3, & \alpha_{i,n-j,m} > 90^\circ \\
    \end{cases}
    ,
    \label{eq:attitude_transition_time}
\end{equation}
\begin{equation}
    \alpha_{i,n-j,m} = |\theta_{i, n} - \theta_{j, m}| + |\phi_{i, n} - \phi_{j, m}| + |\psi_{i, n} - \psi_{j, m}|
    \label{eq:angle_displacement}
    ,
\end{equation}
where $\alpha_{i,n-j,m}$ is the total attitude transition angle between target $i$ observed at $t_{i,n}$ and target $j$ observed at $t_{j,m}$. 

Clouds are modeled as a grid deployed at altitude $h_{clouds}$ above the observation area. Each cell in the grid represents a square area of size $l_{clouds}$ centered at a specific coordinate, and is assigned a binary variable indicating the presence or absence of clouds in that area. $P_{clouds}$ denotes the percentage of cells representing the presence of clouds within the grid. Cells representing cloud presence are grouped into clusters of adjacent cells, with each cluster emulating an individual cloud. $\delta_{i,n}$ represents the fraction of the target's surface covered by clouds when capturing target $i$ at $t_{i,n}$.

Atmospheric turbulence is determined by the refractive index structure parameter $C_n^2$, which is a statistical measure of the strength of the turbulence in the atmosphere, and, similarly to clouds, it is modeled as a grid deployed at altitude $h_{C_n^2}$ above the observation area. Each cell in the grid represents a square area of size $l_{C_n^2}$ centered at a specific coordinate, and is assigned a $C_n^2$ value. $P_{C_{n, \max}^2}$ denotes the percentage of cells with a $C_n^2$ value greater than $C_{n, \max}^2$ in the grid, with $C_{n, \max}^2$ being the maximum allowed $C_n^2$ for obtaining a high-quality image. Since the channel characteristics determining $C_n^2$ values vary by position and are independent, the value of each cell is uncorrelated with the others. $C_n^2(i,n)$ represents the $C_n^2$ value affecting the image quality when capturing target $i$ at $t_{i,n}$. 

Both, $C_n^2(i,n)$ and $\delta_{i, n}$ values determine whether  meteorological conditions for observing target $i$ at $t_{i,n}$ are suitable for capturing the image. This suitability is represented by the following binary function:
\begin{equation}
    f_{i,n}(C_n^2(i,n), \delta_{i, n}) = 
    \begin{cases}
        1, & \text{$C_n^2(i,n) \leq C_{n \max}^2$}\\
        & \text{and $\delta_{i, n} < \delta_{\max}$} \\
        0, & \text{otherwise}\\
    \end{cases}
    ,
    \label{eq:binary_feasibility_eq}
\end{equation}
where $\delta_{\max}$ is the maximum permissible fraction of the target's surface that can be covered by clouds. If $f_{i,n}(C_n^2(i,n), \delta_{i, n}) = 1$, the conditions are deemed suitable; otherwise, $f_{i,n}(C_n^2(i,n), \delta_{i, n}) = 0$.

The \gls{aeossp} to be formulated in this scenario is a dual decision-making problem involving the selection of the sequence of targets and determining the capture time for each. To address this, the system is modeled as a directed graph $G = (V,  E)$, where $V$ represents the set of nodes in the graph and $E$ denotes the set of edges connecting them. Each node $v_{i,n}$ denotes a feasible pair observation target $i$ at a capture time $n$, referred to as an action; and the edges $e_{i,n-j,m}$ connect nodes that can be scheduled in the same sequence while meeting the constraints. The graph also includes a node representing the last action taken, as well as a virtual node at the beginning of the observation which connects to all other nodes.

The information gathered at each node $v_{i,n}$ in the graph includes: the \gls{gsd} of target $i$ at $t_{i, n}$; $f_{i,n}(C_n^2(i,n), \delta_{i, n})$, representing the suitability of  meteorological conditions when capturing target $i$ at $t_{i, n}$; and the capture time $t_{i, n}$. Thus, we form the node features vector $\vec{v}_{i, n} = (GSD_{i, n}, f_{i,n}(C_n^2(i,n), \delta_{i, n}),t_{i, n})$. It is assumed that this information is available and can be processed to create the graph.

\section{Agile Earth Observation Satellite Scheduling Problem} \label{sec:aeossp}

The \gls{aeossp} consists of maximizing the total observation profit while meeting temporal, energy, and memory operational constraints. For the sake of simplicity, we have only considered time constraints. Interestingly, this results in significant energy and memory savings, as an optimal resolution of the \gls{aeossp} leads to a better resource management, as later shown in the results. The mathematical formulation is as follows:
  \vspace{-0.5em}
\begin{subequations}
  \begin{align}
                \text{max} \quad &\sum_{i \in T} \rho_{i,n} x_i, \label{eq:4a}\\
    \text{s.t.} \quad &sw_i \leq t_{i, n} \leq ew_i - d_i,\label{eq:4b}\\
&t_{i, n} + d_i + \Delta t_{i,n-j,m} \leq t_{j, m},\label{eq:4c}\\
                & x_i \in \{0, 1\}, x_i \leq 1,\label{eq:4d}
  \end{align}
  \vspace{-1em}
\label{eq:problem_formulation}
\end{subequations}

\noindent where $\rho_{i,n}$ is the observation profit for target $i$ at $t_{i,n}$; $x_i$ a binary decision variable for target $i$, $x_i = 1$, denotes that target $i$ is scheduled to be observed, otherwise $x_i = 0$; $sw_i$ and $ew_i$ the start and observation time of the \gls{vtw} for target $i$; $d_i$ the duration of the \gls{otw} for target $i$; and $\Delta t_{i,n-j,m}$ the attitude transition time between target $i$ observed at $t_{i,n}$ and target $j$ observed at $t_{j,m}$. Equation (\ref{eq:4a}) is the optimization objective function, which is to maximize the sum of collected profits; Equation (\ref{eq:4b}) represents the \gls{vtw} constraint, which ensures that the observation of target $i$ occurs within its \gls{vtw}; Equation (\ref{eq:4c}) represents the attitude transition constraint, which guarantees that observing target $j$ at $t_{j, m}$ is feasible after observing target $i$ at $t_{i,n}$; Equation (\ref{eq:4d}) represents the value of the decision variable and the fact that a target can exist at most once in a single scheduling.

\begin{figure} [t]
\centering
\includegraphics[width=0.45\textwidth]{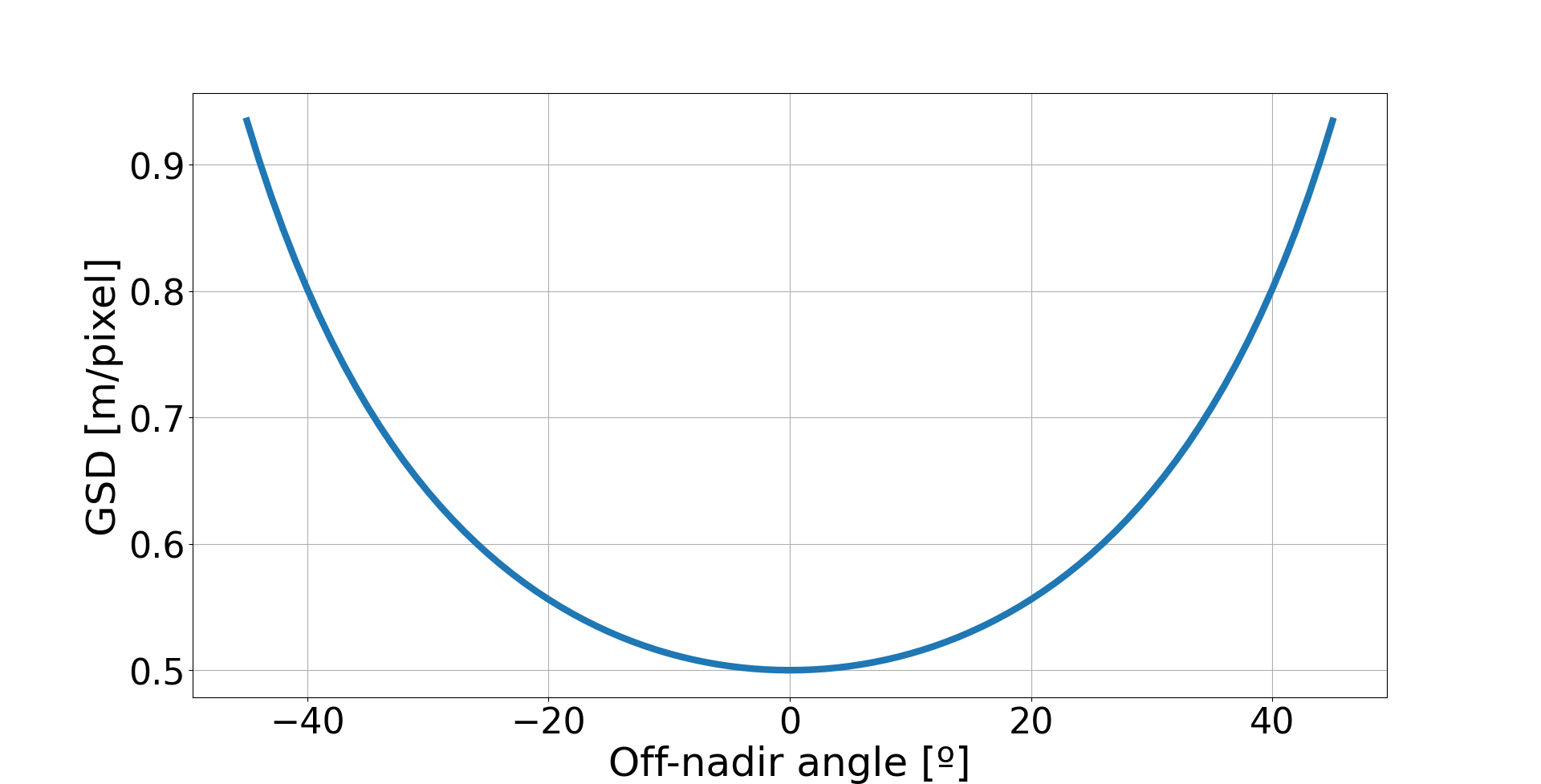}
\caption{Variation of \gls{gsd} with respect to the off-nadir angle, assuming a \gls{gsd} of 0.5 m/pixel at nadir.}
\label{fig:gsd_variation} \vspace{-0.4cm}
\end{figure}

The observation profit $\rho_{i,n}$ ranges from 0 to 1, where 1 indicates that the conditions at $t_{i, n}$ allow us to achieve the highest possible image quality, and 0 denotes that the conditions are not appropriate for capturing the image at $t_{i, n}$. We formulate it as:
\begin{equation}
    \rho_{i, n} = \frac{GSD_{nadir}}{GSD_{i, n}} f_{i, n}(C_n^2(i,n), \delta_{i, n}),
    \label{eq:observation_profit}
\end{equation}
where $GSD_{nadir}$ and $GSD_{i,n}$ are the \gls{gsd} at nadir and for target $i$ at $t_{i,n}$, respectively. As mentioned previously, the \gls{gsd} increases as the target moves away from the nadir reaching its minimum at the nadir itself (see Fig. \ref{fig:gsd_variation}), i.e, the minimum \gls{gsd} for each target is obtained in the middle of its \gls{vtw}. Thus, the first term of the observation profit refers to image resolution. A small \gls{gsd} value leads to a better resolution. The second term, $f_{i, n}(C_n^2(i,n), \delta_{i, n})$, represents the suitability of meteorological conditions.

\section{Algorithm} \label{sec:algorithm}
\begin{comment}
\begin{enumerate}
    \item {\color{blue} Graph formulation}
    \item {\color{blue} Algorithm}
\end{enumerate}  
\end{comment}

The solution of the \gls{aeossp} can be considered a sequential decision process in which the next action is selected based on the current graph, representing the state. Once the decision is made, the graph is updated so it includes only the last selected pair observation target $i$ at a capture time $n$ and the ones that are still feasible. This process is repeated until the resulting graph has only the last taken action, meaning that there are no more feasible actions.

Thus, we define a \gls{mdp} $\langle S, A, T, R \rangle$, where: $S$ is the state set represented by the graph; $A$ the set of possible actions represented by the nodes in the graph; $T$ the state transition function, which corresponds to the graph update based on the last selected action; and $R$ the reward function, which corresponds to the observation profit when it is greater than zero, otherwise, it will correspond to a penalty.

The decisions that compose the final sequence of actions are made by a neural network. We make use of \gls{gat} \cite{veličković2018graphattentionnetworks}, a \gls{gnn} structure that introduces the mechanism of attention to capture the most relevant characteristics of the graph. \gls{gat}s are able to weight the relationships between the nodes of the graph based both on the features of the nodes and the edge structure. 

\begin{figure} [t]
\centering
\includegraphics[width=0.43\textwidth]{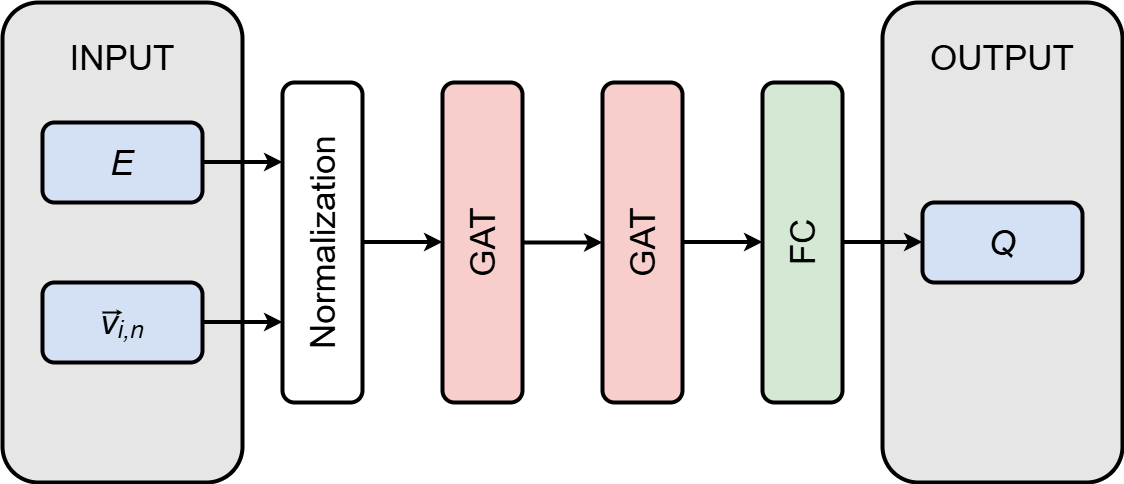}
\caption{\gls{gnn} architecture.}
\label{fig:network_architecture} \vspace{-0.4cm}
\end{figure}

Similar to \cite{math11194059}, we propose a model with two single-layer \gls{gat}s as embedding layers. We transfer the normalized features vector of each node $v_{i,n}$ along with the edges structure $E$ in these layers obtaining the node embeddings. These embeddings, which contain meaningful information based on both node features and edge structure, are then passed through a fully connected layer that outputs a one-dimensional value representing the quality of a particular action. The proposed network architecture is shown in Fig. \ref{fig:network_architecture}.

The parameters of the neural network are obtained by learning from batches of training data following a \gls{dqn} approach. In Q-learning, an agent aims to find the optimal policy that maximizes the long-term cumulative reward. For this, a state-action value $Q_t$ is defined to describe the expected reward $r_t$ of taking action $a_t$ under state $s_t$ at time $t$ so that the agent can learn the optimal policy. Mathematically, this is expressed as:
\begin{align}
    Q_t(s_t, a_t) = & \ (1 - lr)Q_t(s_t, a_t) \notag \\
    & + lr \big(r_t + \gamma \max_a Q_{t+1}(s_{t+1}, a)\big),
    \label{eq:q_learning}
\end{align}
where $\gamma$ is the discount factor that adjust the importance of future rewards over time and $lr$ the learning rate. At each time step the agent takes an action $a_t$, observes the reward $r_t$, and the state is updated to $s_{t+1}$ depending on $s_t$ and $a_t$ until the episode is over. An episode is defined as a complete sequence of interactions between the agent and the environment, starting from an initial state and ending when a specific termination condition is reached. The Q-table is updated with every time step $t$. In a \gls{dqn} framework, a deep neural network is employed to estimate the state-action value $Q_t$. In our context, the agent is the \gls{aeos}; the state $s_t$ the current graph at time $t$; the action $a_t$ the scheduled target based on $s_t$; $r_t$ the observation profit if it is greater than zero, otherwise it corresponds to a penalty; the next state $s_{t+1}$ is the updated graph after selecting the action; the beginning of an episode corresponds to the initial graph with the virtual node and its termination condition is the exhaustion of the set of possible actions. We follow an $\epsilon$-greedy policy: an exploration rate $\epsilon$ is defined, so that the agent choose a random action with a probability $\epsilon$ and the one that maximizes the expected reward with a probability $1 - \epsilon$. We define an initial exploration rate $\epsilon_0$ that decays by a factor $\epsilon_{decay}$ with each episode until a minimum exploration rate $\epsilon_{min}$ is reached in order to favor exploration during the initial episodes. We use mean square error function to calculate the loss and the network parameters are updated according to \gls{sgd}. The training process is shown in \mbox{Algorithm \ref{alg:training}}.

\begin{figure}[t]
\begin{algorithm}[H]
\caption{$\text{DQN}$ algorithm training}
\begin{algorithmic}[1]
\STATE {\bfseries Input:} Set of training episodes $E$, $\epsilon_0$, $\epsilon_{decay}$, $\epsilon_{min}$, $\gamma$, training batch size $\text{batch}_{\text{size}}$
\STATE {\bfseries Output:} {Network parameters}
\STATE Initialize $\epsilon = \epsilon_0$, $\textit{memory} \gets \emptyset$ and $\text{DQN}$
%\STATE Initialize memory of experiences $\textbf{memory} \gets \emptyset$
%\STATE Initialize model $\textbf{DQN}$
%\STATE Define loss function $\textit{Loss}$
\FOR{\text{e} \textbf{in} $E$}
    \STATE Set initial Graph $G$ for episode $e$, $s_t = G$   
    %\STATE $s_t = G$ %\tcp{Initial state}
    \STATE Get initial set of possible actions $A_t$   
    \STATE $\text{done} = False$
    \WHILE{$\text{not done}$}
        \IF{$\text{random(0,1)} > \epsilon$}
            \STATE Sample random action $a_t$ form $A_t$
        \ELSE
            \STATE Sample action $a_t$ from $A_t$ according to $\text{DQN}(s_t, a)$
        \ENDIF
        \STATE Get reward $r_t = R(s_t, a_t)$
        \STATE Update graph $G$, $s_{t + 1} = G$ and set of actions $A_t$
        %\STATE $s_{t + 1} = G$ %\tcp{Next state}
        %\STATE Update set of possible actions $A_t$
        \IF{$A_t$ \text{is empty}}
            \STATE $\text{done} = True$
        \ENDIF
        \STATE Store $\{s_t, a_t, r_t, s_{t + 1}, \text{done}\}$ in \textit{memory}
        \STATE Update state $s_t = s_{t+1}$
        \IF{$\text{length(memory)} \geq  \text{batch}_{\text{size}}$}
        \STATE Sample random training \textit{batch} from $\textit{memory}$
        \FOR{$\{s_t, a_t, r_t, s_{t + 1}, \text{done}\}$ \textbf{in} $\textit{batch}$}
        \STATE $Q_t = \text{DQN}(s_t, a_t)$
        \IF{not done}
            \STATE $Q_{t+1}^{\max} = \argmax(\text{DQN}(s_{t+1}, a))$                    
            \STATE target = $r + \gamma \cdot Q_{t+1}^{\max}$
        \ELSE
            \STATE target = $r$
        \ENDIF
        \STATE \textit{Loss}($Q_t$, $\text{target}$) %\tcp{Calculate losses}
        \STATE Update $\text{DQN}$ using \gls{sgd}
        \ENDFOR
        \ENDIF
    \ENDWHILE
    \IF{$\epsilon > \epsilon_{min}$}
        \STATE $\epsilon = \epsilon \cdot \epsilon_{decay}$ %\tcp{Update $\epsilon$}
    \ENDIF
\ENDFOR
\end{algorithmic}
\label{alg:training}
\end{algorithm}
\vspace{-1.5em}
\end{figure}

\section{Results} \label{sec:results}
%\begin{comment}
\begin{table}[!t]
\caption{Simulation parameters. \label{tab:simulation_parameters}}
\begin{threeparttable}
\begin{tabularx}{\columnwidth}{@{}lX@{}}
\toprule
Parameter & Value \\
\midrule
$d_i$    & $0.15$ s \\
$l_{target}$ & $5$ km \\
$\delta_{\max}$    & $0.25$ \\
$GSD_{nadir}$    & $0.5$ m/pixel \\
$C_{n \max}^2$    & $5 \times 10^{-15}$ $m^{-2/3}$ \\
$\theta_{\max}, \phi_{\max}, \psi_{\max}$ & $45^\circ$, $45^\circ$, $90^\circ$\\
$h_{sat}$ & $600$ km\\
$\zeta$ & $98.6^\circ$\\
$h_{clouds}$ & $10$ km\\
$l_{clouds}$ & $2$ km \\
$P_{clouds}$ & \{0.4, 0.6\} \\
$h_{C_n^2}$ & $20$ km\\
$l_{C_n^2}$ & $10$ km \\
$P_{C_n^2}$ & \{0.2, 0.4\} \\
VTW duration & [18 - 185] s\\
Average observation period & \{1623.79, 977.95\} s\\
$N$ & $\{40, 60, 80, 100\}$ \\
\bottomrule
\end{tabularx}
\end{threeparttable}
\end{table}
%\end{comment}

%\begin{comment}
\begin{table}[!t]
\caption{Learning parameters. \label{tab:learning_parameters}}
\begin{threeparttable}
\begin{tabularx}{\columnwidth}{@{}lX@{}}
\toprule
Parameter & Value \\
\midrule
$\epsilon_0$    & $1$\\
$\epsilon_{decay}$    & $0.999$\\
$\epsilon_{min}$    & $0.01$\\
$\gamma$    & $0.999$\\
Learning rate & $5 \times 10^{-5}$\\
$\text{batch}_{\text{size}}$ & $64$\\
Number of training episodes & $5000$\\
\bottomrule
\end{tabularx}
\end{threeparttable}
\end{table}
%\end{comment}

In order to simplify the problem, we make some assumptions. We assume that all the targets in the set have the same size, that can be captured in a single shot, and that the duration of the observation is the same for each target. We also set only three possible observation times for each target when using the proposed model: at the beginning, in the middle, and at the end of its \gls{vtw}; and an initial satellite attitude of $\theta_{0} = 0^\circ$, $\phi_{0} = 0^\circ$, $\psi_{0} = 0^\circ$. The simulation parameters are shown in Table \ref{tab:simulation_parameters}, while Table \ref{tab:learning_parameters} displays the learning parameters. Code execution and training use an Nvidia A100 GPU with 1 TB of RAM. The algorithm and the simulator are implemented in Python and PyTorch 2.2.0 is used as learning framework.

The performance of the algorithm will be compared with two baselines: (1) \textbf{MaxResolution}, in which the observation time of each target is set to the middle of its \gls{vtw}, i.e., when we get the best possible resolution for the image. The scheduling is decided according to this parameter and meteorological conditions are not taken into account; (2) \textbf{MaxTargets}, in which we aim to capture as many observation targets as possible. These are arranged in ascending order according to their \gls{vtw} and those that are feasible are captured at their earliest observation start time \cite{Pralet_Verfaillie_2013}. Image resolution and meteorological conditions are not taken into account.

To evaluate the performance of both the proposed model and the baselines, four metrics will be used: the total observation profit obtained at the end of the episode, as described in Section \ref{sec:aeossp}, the precision, the percentage of discarded images, and the energy waste due to attitude maneuvers. We define the precision as follows:
\begin{equation}
    \text{Precision} = \frac{N_{sch}'}{N_{sch}},
\end{equation}
being $N_{sch}$ the total number of targets in the final scheduling and $N_{sch}'$ the number of targets in the final scheduling whose observation profit is greater than zero. A higher precision value leads to a more efficient management of memory and energy resources, as it reduces both the number of discarded images and the corresponding attitude maneuvers required to capture them. The energy consumption is quantified in units of energy per unit of attitude transition time \cite{Wang_2021}, so we can calculate the energy waste due to these actions.

\subsection{Training}

\begin{figure} [t]
\centering
\includegraphics[width=0.48\textwidth]{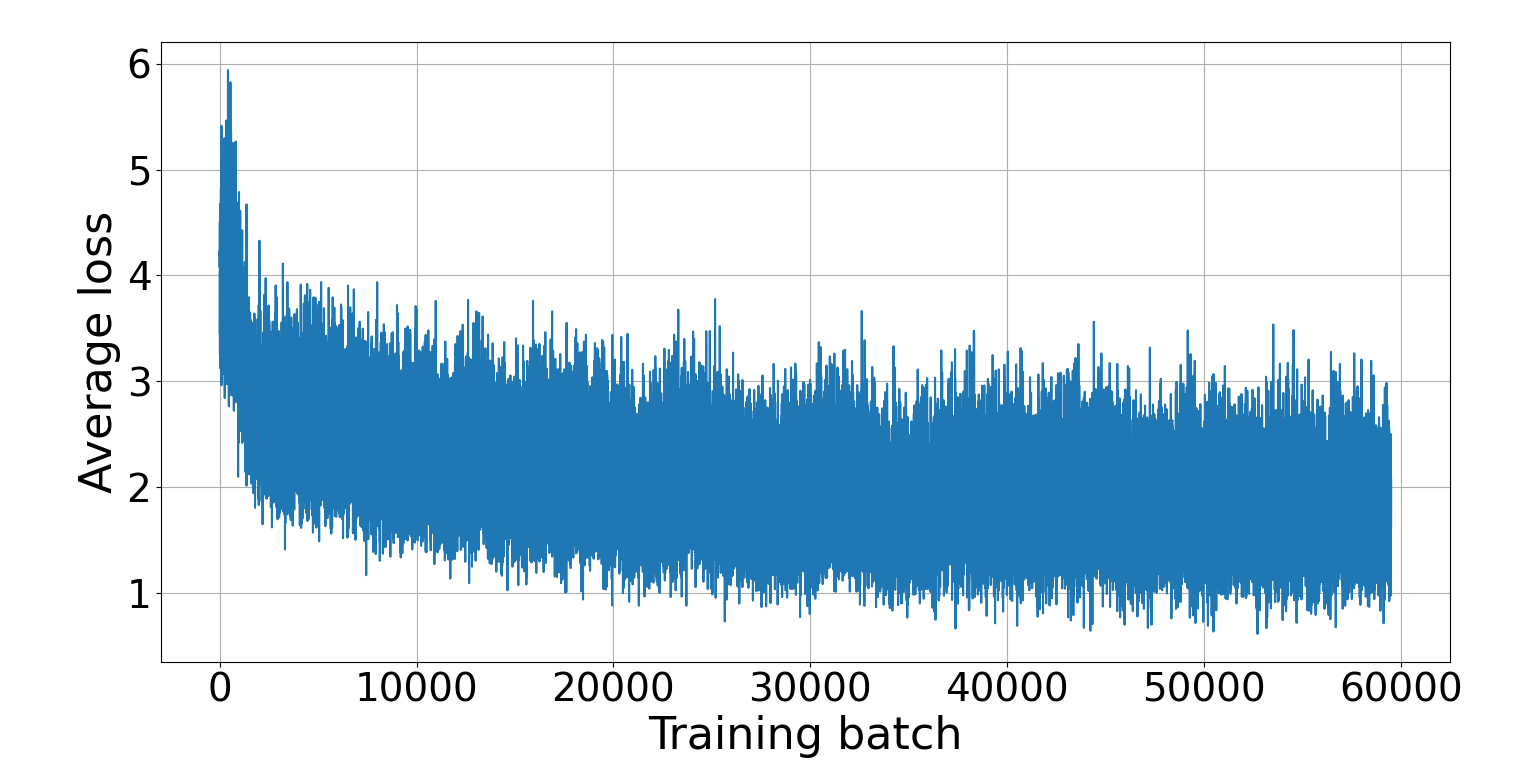}
\caption{Average loss per training batch.}
\label{fig:training} \vspace{-0.4cm}
\end{figure}

\begin{figure} [t]
\centering
\includegraphics[width=0.48\textwidth]{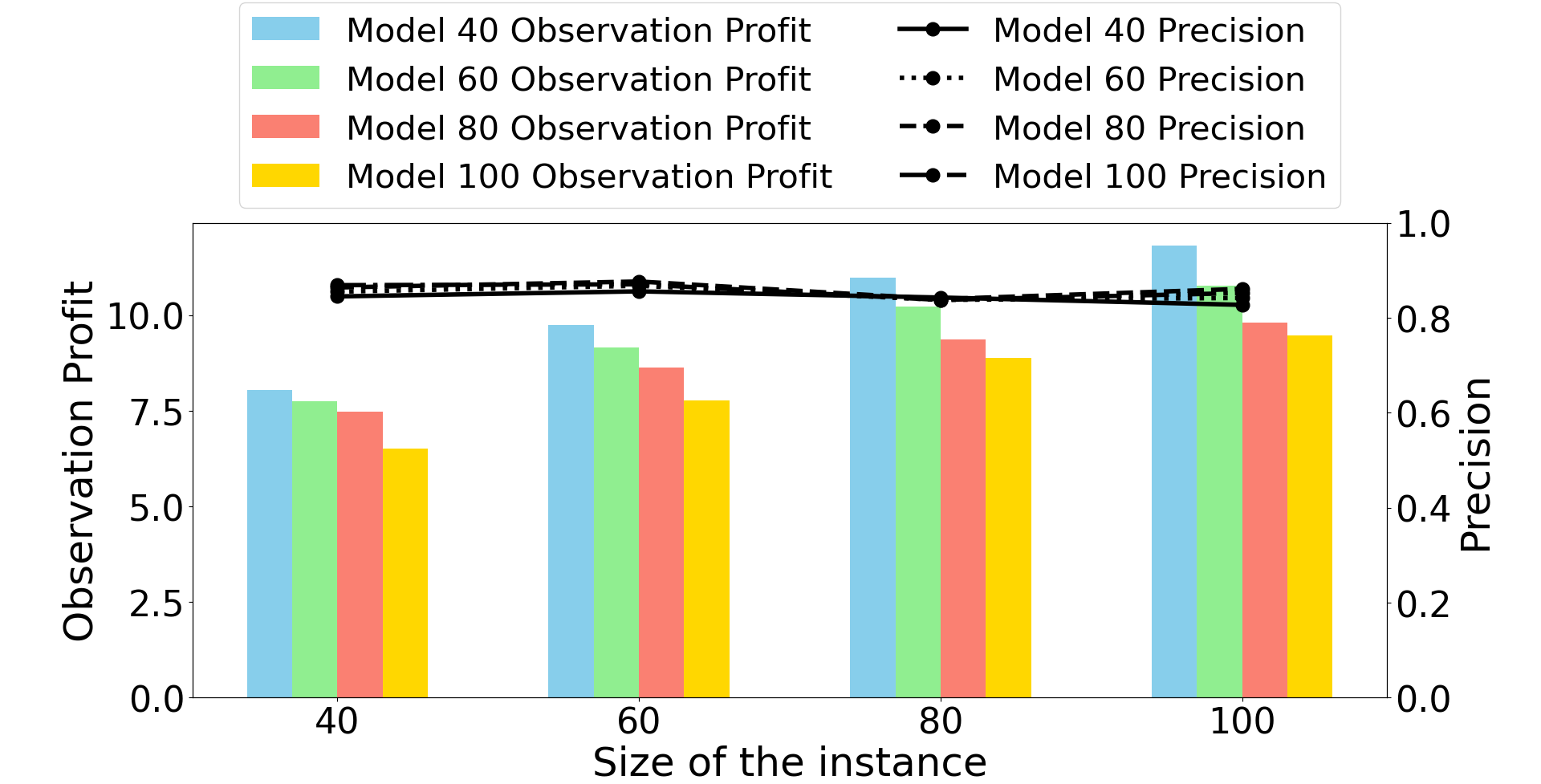}
\caption{Observation profit and precision for different graph sizes ($N = $ 40, 60, 80, 100) and trained model.}
\label{fig:model_comparison} \vspace{-1.5em}
\end{figure}

First, we train the neural network with instances of targets randomly generated of four different instance sizes: $N = $ 40, 60, 80 and 100 targets. Thus, we obtain four different models depending on the size of the training instances. We consider an average observation period, i.e., the period in which the \gls{vtw} of the targets are distributed, of 1623.79 seconds, and clouds and atmospheric turbulence are randomly distributed above the observation area according to $P_{clouds} = 0.4$ and $P_{C_n^2} = 0.2$ for each instance. Fig. \ref{fig:training} shows the average loss per training batch for the case of the training with instances of 40 targets, where a rapid decay and convergence can be observed. We also test the four trained models against instances of size $N = $ 40, 60, 80 and 100 targets and generated in the same conditions as the training (200 instances each). Results are displayed in Fig. \ref{fig:model_comparison}. Although precision does not vary significantly, the model trained with instances of 40 targets gets the highest observation profit in all cases. Therefore, this model will be the one used in the following evaluations.

\subsection{Model Performance}

\begin{figure*}[t]
\centering
\includegraphics[width=1\textwidth]{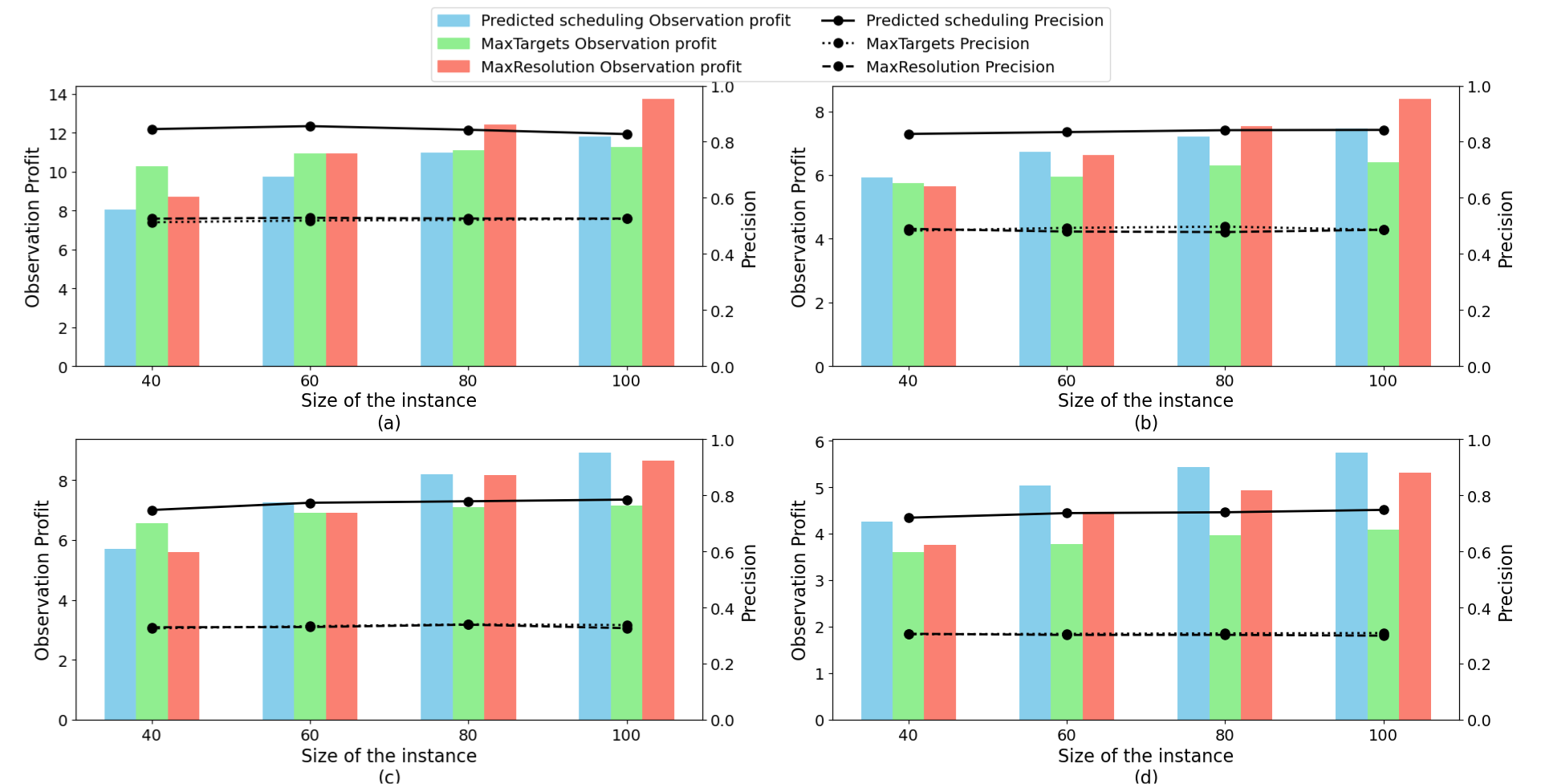}
\caption{Observation profit and precision. (a) Average observation period = 1623.79 seconds, $P_{clouds} = 0.4$, and $P_{C_n^2} = 0.2$. (b) Average observation period = 977.95 seconds, $P_{clouds} = 0.4$, and $P_{C_n^2} = 0.2$. (c) Average observation period = 1623.79 seconds, $P_{clouds} = 0.6$, and $P_{C_n^2} = 0.4$. (d) Average observation period = 977.95 seconds, $P_{clouds} = 0.6$, and $P_{C_n^2} = 0.4$.}
\label{fig:performance_results} \vspace{-0.4cm}
\end{figure*}

We consider four different cases to test the performance of the proposed model and the baselines: (1) instances generated under the same conditions as the training instances; (2) the average observation period is reduced while maintaining the number of targets per instance, so that the decision-making process becomes more critical; (3) more severe meteorological conditions are considered, i.e., higher $P_{clouds}$ and $P_{C_n^2}$ values; (4) a combination of (2) and (3), representing the most challenging case. Like in the model comparison, tests consist of 200 instances.

Fig. \ref{fig:performance_results}a shows the results for the case (1). The observation profit obtained by the model only surpass the MaxTargets baseline in the 100-target instance and ranges from 78\% to 99\% of the baseline values in all other cases, but the precision values are significantly improved, increasing from approximately 52\% to 85\%. This results in a significantly more efficient memory utilization, as the number of discarded images due to cloud coverage or atmospheric turbulence decreases by 68.75\%, and a reduction of a 65\% and 75\% of energy waste from attitude maneuvers with respect to the MaxResolution and MaxTargets baselines is achieved, respectively.

For case (2), we consider an average observation period of 977.95 seconds. The results are displayed in Fig. \ref{fig:performance_results}b. In this case, the observation profit obtained by the predicted scheduling outperforms the one obtained for the baselines except for the 80- and 100-target instances for the MaxResolution baseline. Furthermore the number of discarded images is reduced by 67.65\% and the energy waste by 57\% and 69\% with respect to the MaxResolution and MaxTargets baselines, respectively.

Fig. \ref{fig:performance_results}c illustrates the results for case (3). We consider $P_{clouds} = 0.6$ and $P_{C_n^2} = 0.4$. Similarly to the previous case, the observation profit obtained by the model is greater than the one obtained by the baselines, except in the 40-target instance for the MaxTargets baseline. We also achieve a reduction of 64.2\% in the number of discarded images and 66\% and 78\% of wasted energy with respect to the MaxResolution and MaxTargets baselines, respectively.

Finally, in case (4) we consider an average observation period of 977.95 seconds, $P_{clouds} = 0.6$, and $P_{C_n^2} = 0.4$. Fig. \ref{fig:performance_results}d shows the results. The proposed model achieves a higher observation profit in all cases and reduces the number of discarded images by 61.4\% and the wasted energy by 55\% and by 69\% with respect to the MaxResolution and MaxTargets baselines, respectively.

Thus, we can conclude that the proposed model presents a good management of memory and energy resources. Furthermore, it has a great scalability capacity, since good results are obtained in instances of very different sizes, and outperforms the considered baselines in complex cases, highlighting the importance of an accurate decision making process.

\section{Conclusions} \label{sec:conclusions}

We propose a \gls{drl} approach for the \gls{aeossp} with time-dependent profits considering the impact of the cloud occlusion, atmospheric turbulence and image resolution in order to improve the management of the energy and memory resources. We model the \gls{aeossp} as a directed graph representing the current state that is dynamically updated according to the last action and propose a neural network trained in a \gls{dqn} framework that leverages the potential of the \gls{gat} to decide the best next possible action. The results show that the proposed model reduces the number of discarded images by $> 60\%$ and energy waste from attitude maneuvers by up to $78\%$. Furthermore, it outperforms the baselines considered in challenging cases in terms of observation profit.

Future work will do the integration into the operation of an \gls{eo} satellite network with a joint optimization of the image acquisition process, data processing and communication, enabling more efficient use of the resources of the satellite and the overall system. For this, we will consider a multi-satellite scenario and extend the formulation of the observation profit.

\bibliographystyle{IEEEtran}
\bibliography{refs}

\end{document}